\documentclass[10pt,twocolumn,letterpaper]{article}

\usepackage[pagenumbers]{wacv}

\input{glyphtounicode}
\usepackage{amsmath}
\usepackage{amssymb}
\usepackage{booktabs}
\usepackage{multirow}
\usepackage{graphicx}
\usepackage{xcolor}
\usepackage{enumitem}
\usepackage{algorithm}
\usepackage{algorithmic}
\usepackage{pifont}
\usepackage{placeins}

\newcommand{\method}{MedSTP}
\newcommand{\fullmethod}{Subspace-Transition Token Pruning}
\newcommand{\stdcell}[2]{#1{\tiny\,$\pm$#2}}

\definecolor{wacvblue}{rgb}{0.21,0.49,0.74}
\usepackage[pagebackref,breaklinks,colorlinks,allcolors=wacvblue]{hyperref}

\def\wacvPaperID{1483}
\def\confName{WACV}
\def\confYear{2027}

\title{MedSTP: Preserving Subspace and Transition Evidence for Medical VQA Token Pruning}

\author{%
\parbox{0.96\textwidth}{\centering
Mengchen Fan\textsuperscript{1,*}\quad
Qizhen Lan\textsuperscript{2,*}\quad
Xi Xiao\textsuperscript{1}\\[3pt]
Lin Zhao\textsuperscript{3}\quad
Liang He\textsuperscript{4}\quad
Tianyang Wang\textsuperscript{1}\\[3pt]
Xiaoqian Jiang\textsuperscript{2}\quad
Pulin Che\textsuperscript{1,$\dagger$}\quad
Baocheng Geng\textsuperscript{1,$\dagger$}\\[6pt]
{\normalsize
\textsuperscript{1}University of Alabama at Birmingham\quad
\textsuperscript{2}UTHealth Houston\\[2pt]
\textsuperscript{3}Northeastern University\quad
\textsuperscript{4}ETH Zurich\\[4pt]
\textsuperscript{*}Equal contribution.\quad
\textsuperscript{$\dagger$}Co-corresponding authors.}%
}%
}

\begin{document}
\maketitle

\begin{abstract}

Medical vision-language models (VLMs) for visual question answering (VQA) often pass long visual-token sequences to the language model, making visual-token pruning attractive but risky for medical images, where answers may depend on localized boundary and texture evidence. We study training free pre-decoding pruning and treat pruning as a subspace stable evidence preservation problem. The retained tokens should remain close to the dominant visual evidence subspace induced by the full token set.  To this end, we propose \emph{Subspace-Transition Token Pruning} (\method{}), which estimates each token's spectral support from principal component space and constructs a compact representation subspace. To better reflect medical image structure, \method{} further uses a bounded token-grid transition prior to promote boundary and texture sensitive tokens. We evaluate \method{} on the closed-ended SLAKE and OmniMedVQA benchmarks and on RadImageNet-VQA, which provides both closed-ended and open-ended evaluations, using Lingshu-7B and Qwen3-VL-8B backbones. \method{} delivers consistently strong performance on both the closed-ended and open-ended RadImageNet-VQA evaluations. These results show that \method{} is an effective training free pruning framework for medical domain VQA.
\end{abstract}

\section{Introduction}

Medical VQA is increasingly limited by the visual token budget rather than by the question text. Modern VLMs pass long image-token sequences into the language model, and recent backbones such as Qwen3-VL make this cost central to practical inference~\citep{bai2025qwen3vl}. For medical images, this pressure matters: answers may depend on small boundary or texture changes, not only on the dominant anatomy~\citep{aerts2014radiomics}.

\begin{figure}[t]
\centering
\includegraphics[width=\linewidth]{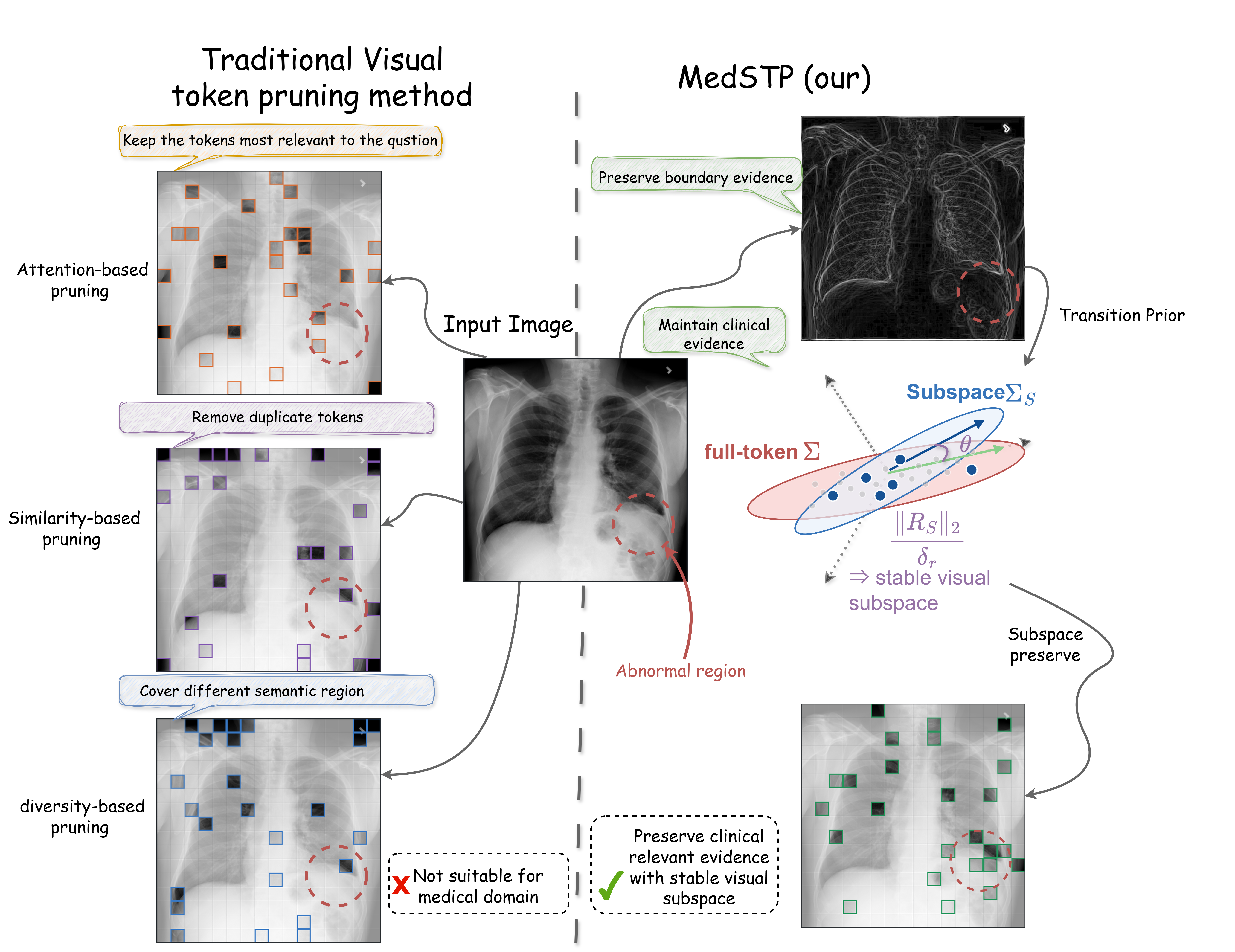}
\caption{Qualitative comparison under the same compact token budget. The dashed red circle marks the relevant abnormal region; \method{} keeps boundary and texture evidence that generic pruning may miss.}
\label{fig:qualitative-token-selection}
\end{figure}

Visual token pruning is the natural compression tool. In ViTs, DynamicViT and EViT remove tokens with learned predictors or class-token attention~\citep{rao2021dynamicvit,liang2022evit}, while Token Merging (ToMe) merges similar patches at runtime~\citep{bolya2023tome}. In VLMs, FastV uses early visual attention~\citep{chen2024fastv}, SparseVLM uses text relevance~\citep{zhang2025sparsevlm}, VisPruner and PyramidDrop target duplicate or low-utility tokens~\citep{zhang2025vispruner,xing2025pyramiddrop}, BTP and PruneSID add diversity or grouping control~\citep{li2025btp,fang2026prunesid}, and TransPrune and Photon use representation transitions or instruction-conditioned scheduling~\citep{li2026transprune,fang2026photon}. This line makes fixed budget pruning practical. It also leaves open what the budget should protect in medical VQA.

The open question is not whether to prune, but what a fixed budget selector should treat as evidence. Most medical token reduction so far lives in segmentation and volume understanding, where dense masks, edge-aware halting, or instruction-conditioned scheduling can exploit task structure during training~\citep{zhou2023tokensparse,guo2025hrvit,fang2026photon}. Frozen-VLM medical VQA is more constrained: before language decoding, the selector must choose \(K\) visual tokens with no dense supervision, no task-specific retraining, and no repeated subset decoding. Under that setting, a selector driven by generic saliency or redundancy can keep the dominant anatomy and still discard boundary or texture tokens that receive lower priority under generic scoring, as illustrated in Fig.~\ref{fig:qualitative-token-selection}.

This failure mode has a representation-level reading. The full token set spans a visual evidence subspace, and the retained subset should stay close enough to it for the language model to receive the leading evidence directions. We take this \emph{subspace-stable evidence preservation} as the selection target. Redundancy then becomes a budget error: a subset may keep high-energy tokens while spending most of its budget on one anatomical pattern. The harder case runs the other way. Subtle boundary and texture transitions can carry subspace support, yet a generic saliency or redundancy score may rank them below the dominant anatomy. Preserving the subspace is necessary, but a rule that scores tokens by global participation alone can still deprioritize these transitions, which can matter in medical images~\citep{aerts2014radiomics}.

We propose \textbf{\fullmethod{} (\method)}, a training free selector for the visual tokens of a frozen VLM. \method{} has two coupled parts. The first forms a subspace-stable subset: token-side PCA defines the visual-evidence subspace, and the retained set is built progressively by selecting, at each step, the token that adds the most previously unrepresented information to the current subspace. A short perturbation argument on the retained-token second moment matrix (Prop.~1) explains why high-participation, non-redundant tokens help keep the leading subspace stable. The second part adds local sensitivity. \method{} reads a Sobel-style transition score from the VLM token grid~\citep{sobel1968isotropic} and applies it as a bounded multiplier on the spectral score, promoting a boundary- or texture-like token when it already carries subspace support. The bound keeps the transition term from overriding the subspace signal.

We make the following contributions:
\begin{itemize}[leftmargin=*]
    \item We cast fixed budget visual-token pruning for frozen-VLM medical VQA as \emph{subspace-stable evidence preservation}: the retained tokens should keep the leading visual subspace induced by the full token set, a target we support with a perturbation argument on the retained-token second moment matrix.
    \item We introduce bounded token-grid transition subspace construction, which promotes boundary and texture tokens when they retain spectral support, without letting the transition term override the subspace signal.
    \item We realize both in \method{}, a training free pre-decoding selector that needs no dense masks or retraining, and evaluate it on SLAKE, OmniMedVQA and RadImageNet-VQA with Lingshu-7B and Qwen3-VL-8B under matched insertion points and fixed token budgets.
\end{itemize}

\section{Related Work}

\subsection{Visual Token Pruning and Subspace Selection}
Compression can act on several different objects. Reweighted optimization compresses DNN weights~\citep{fan2025reweightedcompression}, while federated pruning jointly considers model sparsity and training convergence~\citep{fan2025federatedpruning}. Modality compression instead allocates a limited communication budget across multimodal inputs~\citep{fan2026modalitycompression}. These settings provide a broader context for resource-constrained inference~\citep{lan2026clinicalworkflows,lan2026locationcompression,choi2026d3a}; the object compressed by \method{} is the visual-token sequence of a frozen VLM.

Token reduction has been studied extensively for vision transformers because image patches are often redundant. DynamicViT learns lightweight prediction modules to prune tokens progressively~\citep{rao2021dynamicvit}, EViT reorganizes attentive and inattentive tokens using class token attention~\citep{liang2022evit}, and Token Merging (ToMe) merges similar tokens at runtime without additional training~\citep{bolya2023tome}. These methods establish the basic efficiency trade-off, but their objectives are primarily designed for generic recognition rather than medical VQA.

Recent VLM and MLLM pruning methods further exploit long visual token sequences before or inside language decoding. Attention or relevance based methods score tokens by early layer attention, text-to-visual relation, or instruction guided attention, as in FastV~\citep{chen2024fastv}, SparseVLM~\citep{zhang2025sparsevlm}, and TransPrune~\citep{li2026transprune}. Redundancy based methods remove visually duplicate or low utility tokens, including VisPruner~\citep{zhang2025vispruner}, PyramidDrop~\citep{xing2025pyramiddrop}, and BTP~\citep{li2025btp}. PruneSID combines semantic grouping with intra group non maximum suppression to balance importance and diversity~\citep{fang2026prunesid}. These methods provide strong general VLM pruning baselines, but they are mostly domain agnostic: the retained subset is selected from generic attention, relevance, similarity, or grouping cues, without explicitly modeling the localized boundary, texture, and transition evidence emphasized in medical images.

Recent work has also moved beyond scalar saliency scores. SVD-Prune selects tokens by leverage scores to preserve dominant global variance~\citep{apedo2026svdprune}; ResPrune formulates pruning as text-conditioned subspace reconstruction~\citep{li2026resprune}; and SPARE treats pruning as column subset selection with explicit subspace reconstruction error~\citep{lee2026spare}. 

\method{} draws on this broader subspace motivation but targets a different setting. We study fixed budget pruning before LLM decoding for frozen medical VQA backbones, where dense masks, task-specific retraining, and repeated subset decoding are unavailable. PCA provides a classical tool for identifying dominant low-dimensional subspaces~\citep{abdi2010pca,jolliffe2016pca,fang2026prunesid}. Building on this general subspace view, \method{} uses token-side PCA to define the visual-evidence space. From this PCA representation, the retained medical-evidence subspace is constructed progressively according to the additional subspace information contributed by each candidate, with a bounded token-grid transition prior incorporated into the same selection criterion. This construction allows global representation structure and localized boundary and texture evidence to jointly determine the final retained subset. We evaluate \method{} under matched pre-LLM token budgets and report a subspace-drift diagnostic for the retained subset.

\subsection{Medical VLMs and Medical Token Compression}
Medical VLMs adapt multimodal modeling to biomedical images and clinical questions~\citep{lin2026models,zhu2026medeyes}. MedGemma 1.5 provides a medical multimodal instruction model for image and text understanding~\citep{sellergren2026medgemma15}, while Med-Flamingo adapts Flamingo style few shot multimodal learning to medical image question answering~\citep{moor2023medflamingo}. Lingshu-7B provides a medical MLLM backbone with a Qwen-compatible visual token interface~\citep{lingshu2026modelcard}; Qwen3-VL-8B is our general VLM backbone for testing whether the same selector transfers beyond a medical model~\citep{bai2025qwen3vl}. SLAKE~\citep{liu2021slake} and OmniMedVQA~\citep{hu2024omnimedvqa} provide image question answer pairs across radiology and broader medical modalities. RadImageNet-VQA~\citep{butsanets2026radimagenetvqa} provides CT- and MRI-based radiologic questions in open-ended and closed-ended formats. BrainTRACE extends medical VLM evaluation to longitudinal, multi-sequence, and volumetric evidence in brain MRI reasoning~\citep{lan2026braintrace}, beyond the two-dimensional evaluation studied here.

Medical learning also encompasses population-level time-series analysis, such as studies of vaccine rollout and mental health~\citep{zhai2024vaccinerollout}, and image-based biological inference. In the latter setting, SIGMA transfers fine Golgi morphology cues from high-resolution images to a lower-resolution pathway-state classifier~\citep{fan2026sigma}. For volumetric medical images, Detail Consistent Distillation preserves wavelet-domain detail during teacher--student training for efficient MRI segmentation~\citep{fan2026dcd}. These biomedical applications contextualize the importance of choosing evidence appropriate to a task~\citep{zhu2025pathology}; they do not evaluate visual-token pruning.

Medical token pruning has mainly been studied for image analysis models, segmentation, or volume understanding rather than fixed budget 2D medical VQA. TokenSparse learns score predictors for faster medical image segmentation~\citep{zhou2023tokensparse}; HRViT introduces edge aware token halting to preserve boundary sensitive regions~\citep{guo2025hrvit}; Photon studies instruction conditioned token scheduling for volume understanding~\citep{fang2026photon}; and MedPruner proposes hierarchical token pruning for 3D medical image understanding in VLMs~\citep{liu2026medpruner}. \method{} targets a narrower point in this space: a training free selector that operates directly on frozen VLM visual tokens under a matched pre-LLM token budget for medical VQA.

\subsection{Compact Representations and Knowledge Transfer}
Representation selection and weighting also arise in distributed learning. Gradient matching condenses raw data into representative points~\citep{fan2024representativefusion}, and representative knowledge sharing combines compact datasets with adaptive distillation weights~\citep{casey2025representativeknowledge}. Distributed energy distance measures feature-distribution heterogeneity~\citep{fan2025energydistance}, while PFedDST selects collaborating peers using loss, task similarity, and selection frequency~\citep{fan2025pfeddst}. CRAD weights peer supervision by class-specific reliability~\citep{bilbeisi2026crad}. At the token level, influence calibration allocates on-policy self-distillation supervision according to its relation to the reinforcement-learning objective~\citep{lan2026influencecalibration}. These works address which information to transfer or emphasize during training~\citep{lan2025acamkd,ding2026dprd,zhu2026anatomy}. \method{} instead retains original visual embeddings at inference time, without constructing synthetic training data, selecting peer models, or updating backbone parameters.

\section{Method}

\begin{figure*}[t]
\centering
\includegraphics[width=\textwidth]{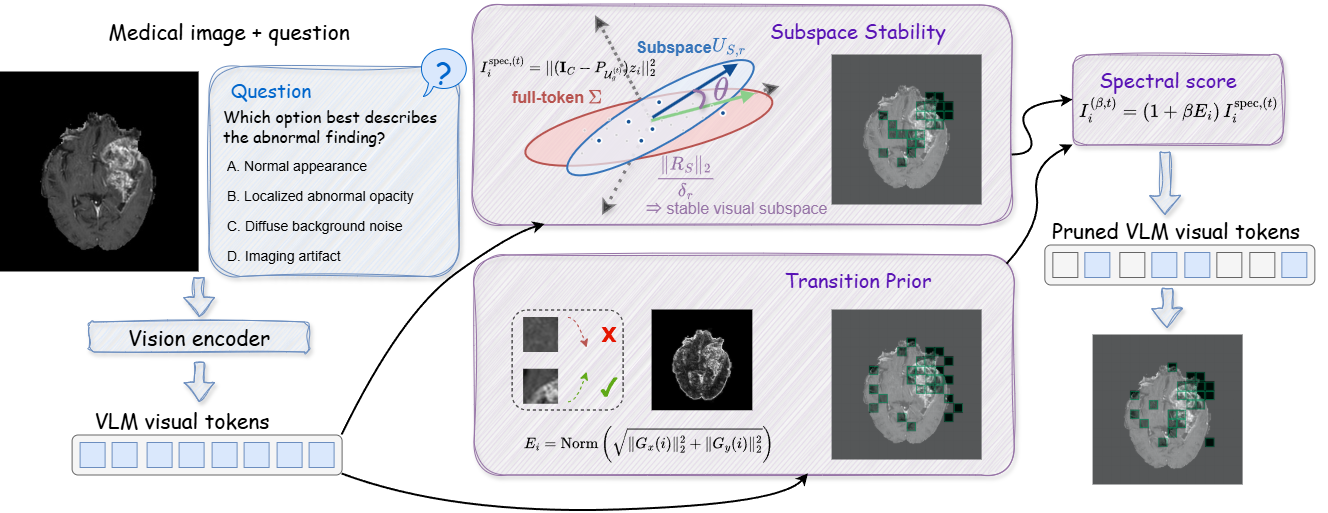}
\caption{Overview of MedSTP for medical VQA token pruning. Given a medical image-question input, the frozen VLM encoder produces visual tokens before decoding. MedSTP constructs a compact retained subspace from principal projection participation and incorporates a bounded token-grid transition multiplier. The selected compact subset is passed to the frozen VQA decoder for answer prediction.}
\label{fig:medstp-overview}
\end{figure*}

\method{} is a training free, pre-decoding selector for the visual tokens passed to a frozen language model. It makes one fixed budget decision from the encoded visual tokens before language decoding, without dense masks, task-specific retraining, or trial decoding of candidate subsets. Figure~\ref{fig:medstp-overview} summarizes the pipeline. Subspace construction keeps tokens that participate in principal evidence directions, while a bounded transition cue incorporates localized medical structure into the same construction process.

\subsection{Problem Setup}

Given an image-question pair \((x,q)\), the visual encoder produces \(N\) tokens:
\begin{equation}
H=\{h_i\}_{i=1}^{N}, \qquad h_i \in \mathbb{R}^{d}.
\end{equation}
For a candidate answer set \(\mathcal{C}(q)=\{a_1,\ldots,a_M\}\), decoding with all visual tokens gives:
\begin{equation}
p_{\mathrm{full}}(a\mid x,q)=p_\theta(a\mid q,H).
\end{equation}
Pruning keeps a subset \(S\subset\{1,\ldots,N\}\) with \(|S|=K\):
\begin{equation}
p_S(a\mid x,q)=p_\theta(a\mid q,H_S).
\end{equation}
An ideal fixed budget selector would keep the answer distribution close to the full-token distribution:
\begin{equation}
S^\star=
\arg\min_{S:\,|S|=K}
D_{\mathrm{KL}}
\left(
p_{\mathrm{full}}(\cdot\mid x,q)
\Vert
p_S(\cdot\mid x,q)
\right).
\label{eq:ideal}
\end{equation}
Equation~\eqref{eq:ideal} states the behavior we would like pruning to preserve. It is not an algorithm; directly optimizing it would require repeated decoding under many candidate subsets, which is outside our inference setting.

The pre-decoding approximation constructs a retained evidence subspace in the principal directions of the full visual token set. A token-grid transition cue then gives a bounded multiplicative boost to supported tokens that sit near local changes in the encoded feature grid. The final subset contains exactly \(K\) visual tokens and is inserted at the same pre-LLM position as the compared pruning baselines.

\subsection{A Subspace Preservation View}

The oracle objective in Eq.~\eqref{eq:ideal} is defined at the answer distribution level, while the selector only observes visual token representations. We therefore use a representation-level target: after pruning, the retained tokens should preserve the leading subspace induced by the full token set. Let \(\alpha_i\ge0\) denote an abstract evidence weight for token \(i\). The full-token weighted second moment matrix is:
\begin{equation}
\Sigma
=
\frac{1}{\sum_{i=1}^{N}\alpha_i+\epsilon}
\sum_{i=1}^{N} \alpha_i h_i h_i^\top
\in \mathbb{R}^{d\times d}.
\label{eq:full-cov}
\end{equation}
The retained subset \(S\) induces the corresponding restricted second moment matrix:
\begin{equation}
\Sigma_S
=
\frac{1}{\sum_{i\in S}\alpha_i+\epsilon}
\sum_{i\in S}\alpha_i h_i h_i^\top .
\label{eq:retained-cov}
\end{equation}
Let \(U_r\) and \(U_{S,r}\) denote the top \(r\) eigenspaces of \(\Sigma\) and \(\Sigma_S\). Define the residual and eigengap as:
\begin{equation}
\delta_r=\lambda_r(\Sigma)-\lambda_{r+1}(\Sigma)>0,
\qquad
R_S=\Sigma_S-\Sigma.
\end{equation}
A natural fixed budget subspace objective is then:
\begin{equation}
\min_{S:\,|S|=K}
\|R_S\|_2.
\label{eq:spectral-objective}
\end{equation}
Its role is to make the design target explicit: a retained set is preferable when its restricted second moment stays close to the full-token second moment. The eigengap \(\delta_r\) is not optimized; it appears in the perturbation view below as the factor that controls how strongly the symmetric residual \(R_S\) can rotate the leading subspace.

\paragraph{Proposition 1 (subspace-branch perturbation bound).}
Assume \(\Sigma\) and \(\Sigma_S\) are symmetric positive semidefinite and \(\|R_S\|_2<\delta_r/2\). Then the leading retained token subspace satisfies:
\begin{equation}
\left\|
\sin\Theta(U_r,U_{S,r})
\right\|_2
\le
\frac{2\|R_S\|_2}{\delta_r}.
\label{eq:dk-bound}
\end{equation}

\emph{Proof sketch.}
Pruning replaces \(\Sigma\) by \(\Sigma_S=\Sigma+R_S\). Weyl's inequality gives spectral separation of at least \(\delta_r/2\) between the retained and discarded eigenspaces when \(\|R_S\|_2<\delta_r/2\). Applying the Davis--Kahan sin theta theorem to \((\Sigma,\Sigma_S)\) yields Eq.~\eqref{eq:dk-bound}~\citep{davis1970rotation}.

Proposition~1 is a perturbation view of the spectral branch. It does not prove that the discrete selector minimizes Eq.~\eqref{eq:spectral-objective}, nor that answer accuracy is preserved. The small-perturbation condition \(\|R_S\|_2<\delta_r/2\) may also fail under aggressive pruning. We use the bound more narrowly: the retained set should preserve the leading evidence structure. The later subspace-drift diagnostic in Fig.~\ref{fig:slake-subspace-drift} checks the realized eigenspace change directly.
\subsection{Subspace Construction}
We instantiate the abstract weight \(\alpha_i\) with a spectral support proxy computed from token-side PCA. Let \(H\in\mathbb{R}^{N\times d}\) be the visual-token matrix. We first apply an element-wise sigmoid transformation to each token feature. Setting \(X=\sigma(H)^\top\in\mathbb{R}^{d\times N}\), PCA on \(X\) gives a token projection matrix \(V\in\mathbb{R}^{N\times C}\) and singular values \(s\in\mathbb{R}^{C}\) over the top \(C\) principal directions. Row \(V_{i,:}\) describes how strongly token \(i\) participates in these directions.
We represent token \(i\) in the resulting subspace as
\(
z_i
=
\operatorname{diag}(s)V_{i,:}^{\top}
\in\mathbb{R}^{C}
\). We define the total spectral evidence carried by token \(i\) as \(\alpha_i = \lVert z_i \rVert_2^2.\) Thus, \(\alpha_i\) provides a practical realization of the abstract evidence weight.

With $P_{ic}=|V_{ic}|$, tokens are assigned to their strongest projection direction, \(g_i=\arg\max_c P_{ic}\). Let \(n_g\) denote the number of tokens assigned to each nonempty group and let \(G\) be the number of nonempty groups. Each group first receives one token, and the remaining \(K-G\) slots are allocated to \(n_g\). Cumulative rounding and a deterministic adjustment ensure that the resulting budgets \(K_g\) satisfy \(\sum_g K_g=K\).

Let \(S_g^{(t)}\) denote the tokens retained from group \(g\) after \(t\) selections. These tokens define the current retained evidence subspace
\begin{equation}
\mathcal U_g^{(t)}
=
\operatorname{span}
\{z_j:j\in S_g^{(t)}\},
\qquad
\mathcal U_g^{(0)}=\{0\}.
\label{eq:retained-evidence-subspace}
\end{equation}

\begin{equation}
\alpha_i
=\left\|
\mathbf P_{\mathcal U_g^{(t)}}
z_i
\right\|_2^2
+
\left\|
\left(
\mathbf I_C-P_{\mathcal U_g^{(t)}}
\right)z_i
\right\|_2^2,
\label{eq:pca-influence0}
\end{equation}

We therefore define the step-dependent spectral contribution
of candidate $i$ as
\begin{equation}
I_i^{\mathrm{spec},(t)}
=
\left\|
\left(
\mathbf I_C-P_{\mathcal U_g^{(t)}}
\right)z_i
\right\|_2^2,
\label{eq:pca-influence}
\end{equation}
where $P_{\mathcal{U}_g^{(t)}}$ is the orthogonal projector
onto the currently retained span and $I_C$ is the
$C$-dimensional identity matrix. A large \(I_i^{\mathrm{spec},(t)}\) indicates that token \(i\)  has a substantial component outside the current subspace. Equation~\eqref{eq:pca-influence} provides the subspace component of the MedSTP selection criterion. The next subsection incorporates localized transition evidence into this quantity and defines the final token-selection and subspace-update rule. This construction is motivated by the preservation objective in Eq.~\eqref{eq:spectral-objective}.

\subsection{Local Transitions under a Spectral Constraint}
The spectral proxy is global: it can preserve principal directions while still ranking subtle boundary or texture transitions below dominant anatomy. To add local sensitivity, \method{} reshapes the \(N\) visual tokens according to the spatial token grid provided by the visual encoder. 

For \(\mathcal{H}\in\mathbb{R}^{H_g\times W_g\times d}\), fixed Sobel kernels~\citep{sobel1968isotropic} are applied channel-wise:
\begin{equation}
S_x=
\begin{bmatrix}
-1 & 0 & 1\\
-2 & 0 & 2\\
-1 & 0 & 1
\end{bmatrix},
\quad
S_y=
\begin{bmatrix}
-1 & -2 & -1\\
0 & 0 & 0\\
1 & 2 & 1
\end{bmatrix}.
\end{equation}
Let \(G_x=S_x*\mathcal{H}\) and \(G_y=S_y*\mathcal{H}\), with the reshaped grid indexed back to the original token id \(i\). The transition magnitude and normalized prior are
\begin{equation}
\begin{aligned}
T_i
&=
\sqrt{
\left\|G_x(i)\right\|_2^2+
\left\|G_y(i)\right\|_2^2+\epsilon
},
\\
E_i
&=
\frac{
T_i-\min_j T_j
}{
\max_j T_j-\min_j T_j+\epsilon
}.
\end{aligned}
\label{eq:sobel-score}
\end{equation}
Here \(E_i\in[0,1]\) is a Sobel-style transition cue on the VLM token embeddings. It is incorporated into the current subspace contribution through a bounded multiplier:
\begin{equation}
I_i^{(\beta,t)}
=
\left(1+\beta E_i\right)
I_i^{\mathrm{spec},(t)},
\qquad \beta\geq0.
\label{eq:fg-score}
\end{equation}

Let \(\mathcal G_g=\{i:g_i=g\}\) denote the set of tokens assigned to group \(g\). At each step \(t\) in group \(g\), we select
\(i_g^{(t)}=\arg\max_{i\in\mathcal G_g\setminus S_g^{(t)}}I_i^{(\beta,t)}\),
update \(S_g^{(t+1)}=S_g^{(t)}\cup\{i_g^{(t)}\}\) and its retained
subspace \(\mathcal U_g^{(t+1)}\), and repeat until \(K_g\) tokens have
been selected. The final retained set is
\(S=\bigcup_{g=1}^{G}S_g^{(K_g)}\).

\section{Experiments}

\subsection{Experimental Setup}

We evaluate \method{} on three medical VQA benchmarks with two VLM backbones: Lingshu-7B~\citep{lingshu2026modelcard}, a medical VLM, and Qwen3-VL-8B~\citep{bai2025qwen3vl}, a general VLM. SLAKE~\citep{liu2021slake} provides 2,394 evaluated image question instances, the \texttt{qt-ai} of OmniMedVQA~\citep{hu2024omnimedvqa} provides 3,329 evaluated instances for anatomy-focused VQA, and RadImageNet-VQA~\citep{butsanets2026radimagenetvqa} provides 1,000 paired Anatomy-MC and Anatomy-open questions over 2D CT and MRI images. This two-backbone setting tests whether the same fixed budget pruning behavior transfers across medical and general VLMs. 

All methods prune visual tokens after the vision language merger and before language model insertion. Semantic correctness of open-ended answers is evaluated independently by two method-blind open-source judges, Selene~\citep{alexandru2025selene} and Phi-4~\citep{abdin2024phi4}, using the same fixed rubric. All results are reported as the point estimate \(\pm\) bootstrap standard error over the evaluated instances. All experiments are run on servers equipped with NVIDIA A100 GPUs. All pruning methods are executed in the same software environment with identical prompts, insertion points, and token budgets.

\paragraph{Compared methods.}
We compare against representative pruning and token-selection baselines under the same frozen-backbone protocol. FastV~\citep{chen2024fastv} and SparseVLM~\citep{zhang2025sparsevlm} represent attention- or question-relevance based selection. VisPruner~\citep{zhang2025vispruner}, BTP~\citep{li2025btp}, and PruneSID~\citep{fang2026prunesid} cover redundancy, diversity, and grouping based selection. HRViT is the medical ViT pruning method~\citep{guo2025hrvit}. We compared adapted versions of SVD-Prune~\citep{apedo2026svdprune}, ResPrune~\citep{li2026resprune}, and SPARE~\citep{lee2026spare}, and the results are provided in the Supplementary.

\subsection{Main Results}
Across the three benchmarks and both backbones, \method{} achieves the strongest aggregate performance among the evaluated pruning methods. Table~\ref{tab:omnimed-main} reports the OmniMedVQA results. With Lingshu-7B, \method{} reaches 70.83\% average accuracy, exceeding the strongest competing method by 3.57\%. With Qwen3-VL-8B, it achieves 58.82\%, improving  by 2.89\%. Table~\ref{tab:slake-main} shows a consistent advantage on SLAKE. \method{} improves the average accuracy over the strongest competing baseline by 1.55\% with Lingshu-7B and by 1.67\% with Qwen3-VL-8B, reaching 76.78\% and 78.45\%, respectively.

Table~\ref{tab:radimagenet-main} further evaluates this transfer on both closed- and open-ended RadImageNet-VQA questions. With Lingshu-7B, \method{} improves by 3.50\% on closed-ended average accuracy and by 5.20\% on the method-blind open-ended judge average. With Qwen3-VL-8B, the corresponding gains are 1.60\% and 1.25\%.

\begin{table*}[t]
\centering
\footnotesize
\setlength{\tabcolsep}{4.1pt}
\caption{OmniMedVQA result at 10\% keep ratio. The OmniMedVQA categories are OS = organ/structure, RJ = radiology/joint/body region, LC = lesion/cellular/tissue morphology, and ES = eye/skin. Avg. is the overall accuracy. All values except KL are percentages.}
\label{tab:omnimed-main}
\begin{tabular}{@{}clcccccc@{}}
\toprule
\textbf{Model} & \textbf{Method} & \textbf{OS Acc.\(\uparrow\)} & \textbf{RJ Acc.\(\uparrow\)} & \textbf{LC Acc.\(\uparrow\)} & \textbf{ES Acc.\(\uparrow\)} & \textbf{Avg. Acc.\(\uparrow\)} & \textbf{KL\(\downarrow\)} \\
\midrule
\multirow{9}{*}{\textbf{Lingshu-7B}}
& Full tokens & \stdcell{73.80}{1.15} & \stdcell{83.90}{1.20} & \stdcell{86.82}{1.19} & \stdcell{95.94}{1.78} & \stdcell{80.62}{0.69} & \stdcell{0.000}{0.000} \\
\cmidrule(lr){2-8}
& Random & \stdcell{22.29}{1.09} & \stdcell{64.62}{1.56} & \stdcell{33.21}{1.66} & \stdcell{69.92}{4.14} & \stdcell{38.69}{0.84} & \stdcell{1.063}{0.017} \\
& BTP & \stdcell{33.26}{1.23} & \stdcell{64.19}{1.56} & \stdcell{52.24}{1.76} & \stdcell{\textbf{89.43}}{2.77} & \stdcell{48.69}{0.87} & \stdcell{0.810}{0.016} \\
& FastV & \stdcell{17.15}{0.99} & \stdcell{56.57}{1.61} & \stdcell{26.62}{1.56} & \stdcell{61.79}{4.38} & \stdcell{32.26}{0.81} & \stdcell{1.103}{0.016} \\
& SparseVLM & \stdcell{18.18}{1.01} & \stdcell{57.63}{1.61} & \stdcell{26.74}{1.56} & \stdcell{62.60}{4.36} & \stdcell{33.07}{0.82} & \stdcell{1.083}{0.016} \\
& VisPruner & \stdcell{45.13}{1.30} & \stdcell{72.88}{1.45} & \stdcell{54.98}{1.75} & \stdcell{79.67}{3.63} & \stdcell{56.65}{0.86} & \stdcell{0.659}{0.015} \\
& HRViT & \stdcell{42.11}{1.29} & \stdcell{68.33}{1.51} & \stdcell{52.36}{1.76} & \stdcell{67.48}{4.22} & \stdcell{52.96}{0.87} & \stdcell{0.703}{0.015} \\
& PruneSID & \stdcell{\underline{61.97}}{1.23} & \stdcell{\underline{79.56}}{1.31} & \stdcell{\underline{62.57}}{1.84} & \stdcell{66.67}{4.25} & \stdcell{\underline{67.26}}{0.80} & \stdcell{\underline{0.390}}{0.012} \\
\cmidrule(lr){2-8}
& \textbf{\method{} (ours)} & \stdcell{\textbf{64.54}}{1.21}
& \stdcell{\textbf{80.19}}{1.30}
& \stdcell{\textbf{70.20}}{1.72}
& \stdcell{\underline{82.11}}{3.46}
& \stdcell{\textbf{70.83}}{0.79}
& \stdcell{\textbf{0.298}}{0.011} \\
\midrule
\multirow{9}{*}{\textbf{Qwen3-VL-8B}}
& Full tokens & \stdcell{56.24}{1.30} & \stdcell{84.43}{1.18} & \stdcell{70.52}{1.61} & \stdcell{69.11}{4.17} & \stdcell{68.16}{0.81} & \stdcell{0.000}{0.000} \\
\cmidrule(lr){2-8}
& Random & \stdcell{34.84}{1.25} & \stdcell{66.53}{1.54} & \stdcell{55.35}{1.75} & \stdcell{65.85}{4.28} & \stdcell{49.92}{0.87} & \stdcell{1.441}{0.039} \\
& BTP & \stdcell{40.05}{1.28} & \stdcell{\textbf{74.89}}{1.41} & \stdcell{\underline{60.82}}{1.72} & \stdcell{66.67}{4.25} & \stdcell{\underline{55.93}}{0.86} & \stdcell{0.942}{0.032} \\
& FastV & \stdcell{38.75}{1.28} & \stdcell{66.00}{1.54} & \stdcell{55.10}{1.75} & \stdcell{\underline{67.48}}{4.22} & \stdcell{51.49}{0.87} & \stdcell{1.114}{0.033} \\
& SparseVLM & \stdcell{35.05}{1.25} & \stdcell{71.50}{1.47} & \stdcell{60.32}{1.73} & \stdcell{\textbf{68.29}}{4.20} & \stdcell{52.72}{0.87} & \stdcell{1.237}{0.038} \\
& VisPruner & \stdcell{40.19}{1.28} & \stdcell{73.94}{1.43} & \stdcell{57.34}{1.74} & \stdcell{65.85}{4.28} & \stdcell{54.85}{0.86} & \stdcell{1.018}{0.033} \\
& HRViT & \stdcell{39.92}{1.28} & \stdcell{71.82}{1.46} & \stdcell{54.48}{1.76} & \stdcell{65.85}{4.28} & \stdcell{53.44}{0.86} & \stdcell{\underline{0.937}}{0.031} \\
& PruneSID & \stdcell{\underline{45.69}}{1.26} & \stdcell{64.30}{1.56} & \stdcell{55.79}{1.87} & \stdcell{66.67}{4.26} & \stdcell{53.89}{0.87} & \stdcell{1.079}{0.035} \\
\cmidrule(lr){2-8}
& \textbf{\method{} (ours)}  & \stdcell{\textbf{52.57}}{1.27}
& \stdcell{66.42}{1.54}
& \stdcell{\textbf{61.02}}{1.83}
& \stdcell{66.67}{4.25}
& \stdcell{\textbf{58.82}}{0.85}
& \stdcell{\textbf{0.689}}{0.026} \\
\bottomrule
\end{tabular}
\end{table*}

\begin{table*}[t]
\centering
\footnotesize
\setlength{\tabcolsep}{4.1pt}
\caption{SLAKE result with official question group breakdown at 10\% keep ratio. The SLAKE categories are OR = organ, AB = abnormality, PO = position, and OT = other.}
\label{tab:slake-main}
\begin{tabular}{@{}clcccccc@{}}
\toprule
\textbf{Model} & \textbf{Method} & \textbf{OR Acc.\(\uparrow\)} & \textbf{AB Acc.\(\uparrow\)} & \textbf{PO Acc.\(\uparrow\)} & \textbf{OT Acc.\(\uparrow\)} & \textbf{Avg. Acc.\(\uparrow\)} & \textbf{KL\(\downarrow\)} \\
\midrule
\multirow{9}{*}{\textbf{Lingshu-7B}}
& Full tokens & \stdcell{88.40}{0.95} & \stdcell{72.78}{1.74} & \stdcell{94.81}{2.53} & \stdcell{86.43}{1.51} & \stdcell{83.92}{0.75} & \stdcell{0.000}{0.000} \\
\cmidrule(lr){2-8}
& Random & \stdcell{65.65}{1.40} & \stdcell{25.84}{1.71} & \stdcell{70.13}{5.22} & \stdcell{75.00}{1.91} & \stdcell{56.93}{1.01} & \stdcell{0.641}{0.017} \\
& BTP & \stdcell{\underline{80.12}}{1.18} & \stdcell{\textbf{59.94}}{1.92} & \stdcell{\textbf{93.51}}{2.81} & \stdcell{81.01}{1.73} & \stdcell{\underline{75.23}}{0.88} & \stdcell{0.190}{0.008} \\
& FastV & \stdcell{64.60}{1.41} & \stdcell{26.76}{1.73} & \stdcell{62.34}{5.52} & \stdcell{68.41}{2.05} & \stdcell{55.01}{1.02} & \stdcell{0.611}{0.015} \\
& SparseVLM & \stdcell{65.30}{1.41} & \stdcell{26.45}{1.72} & \stdcell{61.04}{5.56} & \stdcell{70.35}{2.01} & \stdcell{55.64}{1.02} & \stdcell{0.610}{0.015} \\
& VisPruner & \stdcell{74.63}{1.28} & \stdcell{47.71}{1.95} & \stdcell{87.01}{3.83} & \stdcell{77.91}{1.83} & \stdcell{68.38}{0.95} & \stdcell{0.337}{0.012} \\
& HRViT & \stdcell{70.62}{1.34} & \stdcell{30.58}{1.80} & \stdcell{87.01}{3.83} & \stdcell{72.48}{1.97} & \stdcell{60.61}{1.00} & \stdcell{0.473}{0.014} \\
& PruneSID & \stdcell{79.51}{1.17} & \stdcell{56.12}{1.94} & \stdcell{92.21}{3.07} & \stdcell{\underline{83.14}}{1.67} & \stdcell{74.31}{0.90} & \stdcell{0.221}{0.011} \\
\cmidrule(lr){2-8}
& \textbf{\method{} (ours)} & \stdcell{\textbf{81.78}}{1.14}
& \stdcell{\underline{57.34}}{1.93}
& \stdcell{\textbf{93.51}}{2.81}
& \stdcell{\textbf{87.79}}{1.44}
& \stdcell{\textbf{76.78}}{0.86}
& \stdcell{\textbf{0.169}}{0.009} \\
\midrule
\multirow{9}{*}{\textbf{Qwen3-VL-8B}}
& Full tokens & \stdcell{85.61}{1.04} & \stdcell{65.75}{1.86} & \stdcell{94.81}{2.53} & \stdcell{83.72}{1.63} & \stdcell{80.08}{0.82} & \stdcell{0.000}{0.000} \\
\cmidrule(lr){2-8}
& Random & \stdcell{47.78}{1.47} & \stdcell{48.93}{1.95} & \stdcell{57.14}{5.64} & \stdcell{51.74}{2.20} & \stdcell{49.25}{1.02} & \stdcell{1.824}{0.053} \\
& BTP & \stdcell{\underline{78.90}}{1.20} & \stdcell{\textbf{71.10}}{1.77} & \stdcell{75.32}{4.91} & \stdcell{78.10}{1.82} & \stdcell{76.48}{0.87} & \stdcell{0.455}{0.023} \\
& FastV & \stdcell{65.39}{1.40} & \stdcell{60.86}{1.91} & \stdcell{71.43}{5.15} & \stdcell{77.33}{1.84} & \stdcell{66.92}{0.96} & \stdcell{0.706}{0.027} \\
& SparseVLM & \stdcell{76.55}{1.25} & \stdcell{62.54}{1.89} & \stdcell{70.13}{5.22} & \stdcell{79.84}{1.77} & \stdcell{73.22}{0.90} & \stdcell{0.493}{0.024} \\
& VisPruner & \stdcell{78.73}{1.21} & \stdcell{69.88}{1.79} & \stdcell{\underline{80.52}}{4.51} & \stdcell{\underline{80.62}}{1.74} & \stdcell{\underline{76.78}}{0.86} & \stdcell{\underline{0.401}}{0.023} \\
& HRViT & \stdcell{74.80}{1.28} & \stdcell{64.37}{1.87} & \stdcell{72.73}{5.08} & \stdcell{80.04}{1.76} & \stdcell{73.02}{0.91} & \stdcell{0.514}{0.026} \\
& PruneSID & \stdcell{77.59}{1.23} & \stdcell{69.11}{1.79} & \stdcell{76.62}{4.77} & \stdcell{79.26}{1.78} & \stdcell{75.61}{0.88} & \stdcell{0.416}{0.022} \\
\cmidrule(lr){2-8}
& \textbf{\method{} (ours)} & 
\stdcell{\textbf{81.43}}{1.15}
& \stdcell{\underline{70.64}}{1.78}
& \stdcell{\textbf{81.82}}{4.40}
& \stdcell{\textbf{81.20}}{1.72}
& \stdcell{\textbf{78.45}}{0.84}
& \stdcell{\textbf{0.270}}{0.017} \\
\bottomrule
\end{tabular}
\end{table*}

\begin{table*}[t]
\centering
\scriptsize
\setlength{\tabcolsep}{3.2pt}
\caption{RadImageNet-VQA results at a 10\% visual-token keep ratio. Selene and Phi-4 are method-blind semantic judges.  }
\label{tab:radimagenet-main}
\begin{tabular}{@{}clccccccc@{}}
\toprule
\multirow{2}{*}{\textbf{Model}}
& \multirow{2}{*}{\textbf{Method}}
& \multicolumn{4}{c}{\textbf{Closed-ended}}
& \multicolumn{3}{c}{\textbf{Open-ended}} \\
\cmidrule(lr){3-6}\cmidrule(lr){7-9}
&
& \textbf{CT Acc.$\uparrow$}
& \textbf{MRI Acc.$\uparrow$}
& \textbf{Avg. Acc.$\uparrow$}
& \textbf{KL$\downarrow$}
& \textbf{Selene$\uparrow$}
& \textbf{Phi-4$\uparrow$}
& \textbf{Avg.$\uparrow$} \\
\midrule

\multirow{9}{*}{\textbf{Lingshu-7B}}
& Full tokens
& \stdcell{82.16}{1.85}
& \stdcell{94.77}{0.93}
& \stdcell{89.40}{0.97}
& \stdcell{0.000}{0.000}
& \stdcell{72.40}{1.41}
& \stdcell{68.10}{1.47}
& \stdcell{70.25}{1.39} \\
\cmidrule(lr){2-9}

& Random
& \stdcell{47.42}{2.41}
& \stdcell{63.76}{2.01}
& \stdcell{55.59}{1.57}
& \stdcell{0.866}{0.029}
& \stdcell{44.20}{1.57}
& \stdcell{41.90}{1.56}
& \stdcell{43.05}{1.54} \\

& BTP
& \stdcell{68.31}{2.25}
& \stdcell{71.95}{1.88}
& \stdcell{70.40}{1.44}
& \stdcell{0.543}{0.024}
& \stdcell{49.40}{1.58}
& \stdcell{48.10}{1.58}
& \stdcell{48.75}{1.55} \\

& FastV
& \stdcell{42.96}{2.40}
& \stdcell{33.45}{1.97}
& \stdcell{37.50}{1.53}
& \stdcell{1.247}{0.024}
& \stdcell{29.10}{1.44}
& \stdcell{28.30}{1.43}
& \stdcell{28.70}{1.41} \\

& SparseVLM
& \stdcell{43.90}{2.40}
& \stdcell{34.15}{1.98}
& \stdcell{38.30}{1.54}
& \stdcell{1.236}{0.024}
& \stdcell{28.40}{1.43}
& \stdcell{28.00}{1.42}
& \stdcell{28.20}{1.40} \\

& VisPruner
& \stdcell{64.79}{2.31}
& \stdcell{71.95}{1.88}
& \stdcell{68.90}{1.46}
& \stdcell{0.555}{0.025}
& \stdcell{48.30}{1.58}
& \stdcell{47.60}{1.58}
& \stdcell{47.95}{1.56} \\

& HRViT
& \stdcell{63.85}{2.33}
& \stdcell{74.22}{1.83}
& \stdcell{69.80}{1.45}
& \stdcell{0.530}{0.024}
& \stdcell{48.40}{1.58}
& \stdcell{48.20}{1.58}
& \stdcell{48.30}{1.56} \\

& PruneSID
& \stdcell{\underline{77.70}}{2.02}
& \stdcell{79.97}{1.67}
& \stdcell{\underline{79.00}}{1.29}
& \stdcell{0.348}{0.022}
& \stdcell{\underline{58.90}}{1.56}
& \stdcell{\underline{57.90}}{1.56}
& \stdcell{\underline{58.40}}{1.54} \\
\cmidrule(lr){2-9}

& \textbf{\method{} (ours)}
& \stdcell{\textbf{79.81}}{1.94}
& \stdcell{\textbf{84.49}}{1.51}
& \stdcell{\textbf{82.50}}{1.20}
& \stdcell{\textbf{0.240}}{0.018}
& \stdcell{\textbf{64.80}}{1.51}
& \stdcell{\textbf{62.40}}{1.53}
& \stdcell{\textbf{63.60}}{1.49}\\

\midrule

\multirow{9}{*}{\textbf{Qwen3-VL-8B}}
& Full tokens
& \stdcell{85.45}{1.71}
& \stdcell{92.86}{1.07}
& \stdcell{89.70}{0.96}
& \stdcell{0.000}{0.000}
& \stdcell{76.10}{1.35}
& \stdcell{71.20}{1.43}
& \stdcell{73.65}{1.34} \\
\cmidrule(lr){2-9}

& Random
& \stdcell{61.74}{2.37}
& \stdcell{66.90}{1.97}
& \stdcell{64.32}{1.55}
& \stdcell{1.402}{0.080}
& \stdcell{49.10}{1.61}
& \stdcell{45.00}{1.59}
& \stdcell{47.05}{1.56} \\

& BTP
& \stdcell{81.22}{1.89}
& \stdcell{\underline{89.90}}{1.26}
& \stdcell{86.20}{1.09}
& \stdcell{0.368}{0.042}
& \stdcell{70.60}{1.44}
& \stdcell{66.90}{1.49}
& \stdcell{68.75}{1.42} \\

& FastV
& \stdcell{67.37}{2.27}
& \stdcell{40.77}{2.05}
& \stdcell{52.10}{1.58}
& \stdcell{1.543}{0.063}
& \stdcell{37.90}{1.53}
& \stdcell{37.50}{1.53}
& \stdcell{37.70}{1.53} \\

& SparseVLM
& \stdcell{79.58}{1.95}
& \stdcell{77.53}{1.74}
& \stdcell{78.40}{1.30}
& \stdcell{0.678}{0.051}
& \stdcell{56.30}{1.57}
& \stdcell{54.30}{1.58}
& \stdcell{55.30}{1.56} \\

& VisPruner
& \stdcell{77.70}{2.02}
& \stdcell{84.67}{1.50}
& \stdcell{81.70}{1.22}
& \stdcell{0.501}{0.048}
& \stdcell{66.60}{1.49}
& \stdcell{62.50}{1.53}
& \stdcell{64.55}{1.47} \\

& HRViT
& \stdcell{76.29}{2.06}
& \stdcell{85.19}{1.48}
& \stdcell{81.40}{1.23}
& \stdcell{0.641}{0.060}
& \stdcell{66.80}{1.49}
& \stdcell{62.90}{1.53}
& \stdcell{64.85}{1.46} \\

& PruneSID
& \stdcell{\underline{81.69}}{1.87}
& \stdcell{89.72}{1.27}
& \stdcell{\underline{86.30}}{1.09}
& \stdcell{\underline{0.325}}{0.037}
& \stdcell{\underline{72.40}}{1.41}
& \stdcell{\underline{69.20}}{1.46}
& \stdcell{\underline{70.80}}{1.40} \\
\cmidrule(lr){2-9}

& \textbf{\method{} (ours)}
& \stdcell{\textbf{83.10}}{1.82}
& \stdcell{\textbf{91.46}}{1.17}
& \stdcell{\textbf{87.90}}{1.03}
& \stdcell{\textbf{0.240}}{0.034}
& \stdcell{\textbf{74.20}}{1.38}
& \stdcell{\textbf{69.90}}{1.45}
& \stdcell{\textbf{72.05}}{1.38} \\

\bottomrule
\end{tabular}
\end{table*}

\subsection{Subspace Preservation Diagnostic}

\begin{figure}[t]
\centering
\includegraphics[width=0.8\linewidth]{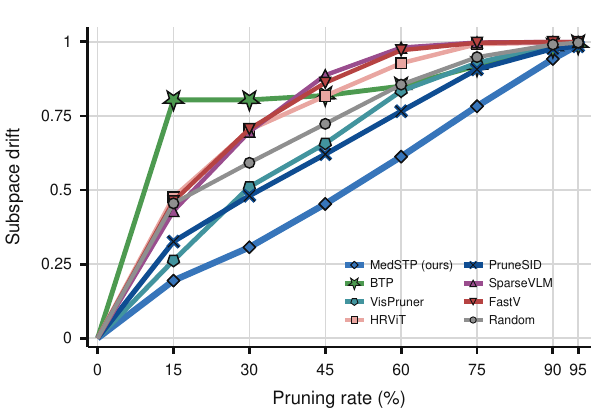}
\caption{Across most pruning rates, \method{} keeps the retained-token eigenspace closer to the full-token visual eigenspace than the competing pruners. Subspace preservation diagnostic on SLAKE with Lingshu-7B; lower drift is better, and 0\% pruning is the zero-drift identity case.}
\label{fig:slake-subspace-drift}
\end{figure}

To examine whether the retained visual tokens preserve the representation structure assumed by the spectral motivation, we compute a subspace drift diagnostic on 200 SLAKE examples with Lingshu-7B. We include 0\% pruning as the identity case with zero drift, as shown in Figure~\ref{fig:slake-subspace-drift}.

\begin{figure}[t]
\centering
\includegraphics[width=0.8\linewidth]{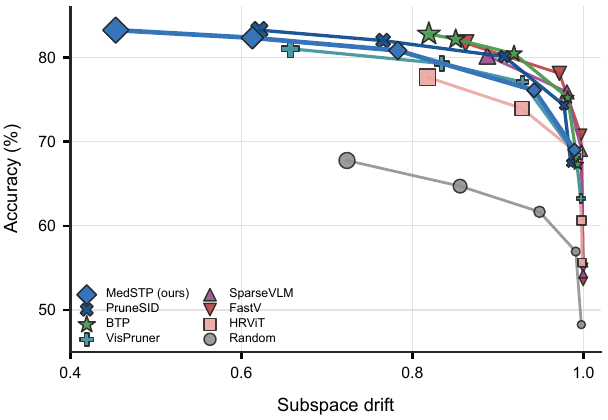}
\caption{As retained-token budgets shrink, \method{} stays nearer the low-drift, high-accuracy corner than the competing pruners. Subspace drift versus answer accuracy on SLAKE with Lingshu-7B; accuracy is reported as a percentage, and larger markers indicate larger retained-token budgets.}
\label{fig:subspace-drift-accuracy}
\end{figure}

The diagnostic curve in Figure~\ref{fig:slake-subspace-drift} shows that \method{} keeps the retained-token eigenspace more closely aligned with the full-token visual eigenspace over most of the evaluated pruning rates. This is the behavior expected from the subspace branch: the language model receives a compact sequence that still tracks the dominant visual directions. Figure~\ref{fig:subspace-drift-accuracy} complements this rate-wise view by plotting realized subspace drift against answer accuracy for each pruning trajectory. As retained-token budgets shrink, methods generally move toward higher drift and lower accuracy. \method{} remains closer to the low-drift, high-accuracy region, suggesting that its retained subset better preserves the representation structure used by the full-token model under aggressive pruning.

\section{Ablation Studies}

\subsection{Accuracy Across Pruning Rates}

\begin{figure*}[t]
\centering
\begin{minipage}[t]{0.45\textwidth}
\centering
\includegraphics[width=0.8\linewidth]{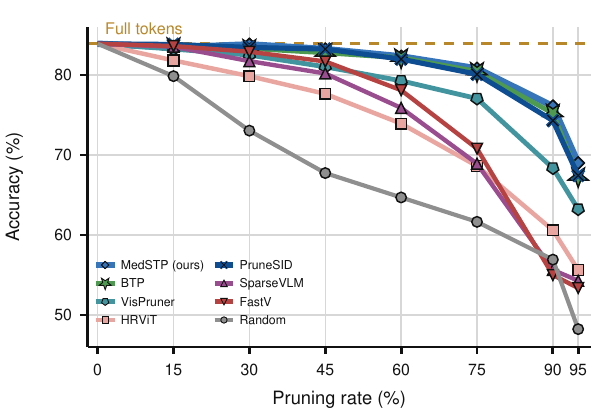}
\vspace{-0.6em}
\centerline{\small (a) Ablation on SLAKE}
\end{minipage}\hfill
\begin{minipage}[t]{0.45\textwidth}
\centering
\includegraphics[width=0.8\linewidth]{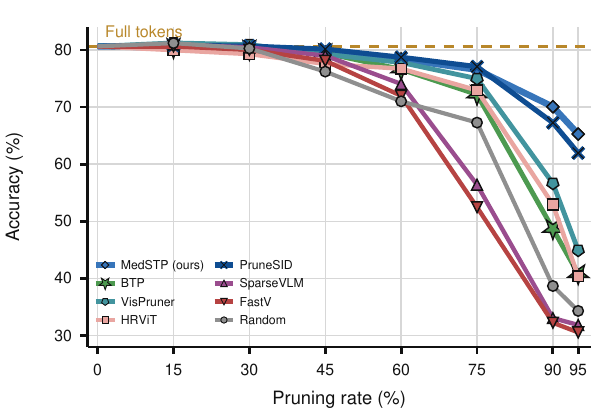}
\vspace{-0.6em}
\centerline{\small (b) Ablation on OmniMedVQA}
\end{minipage}
\caption{VQA accuracy under increasing pruning strength with Lingshu-7B. All methods use matched pruning insertion points and fixed visual-token budgets. Dashed horizontal lines denote full-token accuracy.}
\label{fig:prune-rate-accuracy}
\label{fig:slake-prune-rate}
\label{fig:omnimed-prune-rate}
\end{figure*}

Figure~\ref{fig:prune-rate-accuracy} evaluates how accuracy changes as the pruning rate increases. The curves show the expected tradeoff: most methods remain close to the full token reference under mild compression, but separate more clearly under aggressive pruning. \method{} degrades more gradually on both SLAKE and OmniMedVQA, indicating that its selected subset keeps task relevant visual evidence over a wider range of token budgets.

On SLAKE, the curves stay close under mild pruning and separate mainly at aggressive budgets, matching the main-table pattern that coarse evidence is often enough. OmniMedVQA separates earlier, and \method{} keeps a larger margin in the mid-to-high pruning range, consistent with its greater dependence on localized anatomical and tissue evidence.

\subsection{Transition Prior Ablation}

\begin{table*}[t]
\centering
\scriptsize
\setlength{\tabcolsep}{3.8pt}
\caption{Transition prior ablation with Lingshu-7B at 10\% keep ratio.}
\label{tab:frequency-ablation}
\begin{tabular}{@{}clcccccc@{}}
\toprule
\textbf{Dataset} & \textbf{Variant} & \textbf{OR Acc.\(\uparrow\)} & \textbf{AB Acc.\(\uparrow\)} & \textbf{PO Acc.\(\uparrow\)} & \textbf{OT Acc.\(\uparrow\)} & \textbf{Avg. Acc.\(\uparrow\)} & \textbf{KL\(\downarrow\)} \\
\midrule
\multirow{2}{*}{\textbf{SLAKE}} & w/o trans. & \stdcell{80.99}{1.16}
& \stdcell{55.96}{1.94}
& \stdcell{\textbf{94.81}}{2.53}
& \stdcell{84.50}{1.59}
& \stdcell{75.36}{0.88}
& \stdcell{0.203}{0.010} \\
\cmidrule(lr){2-8}
& \textbf{\method{} (ours)} & \stdcell{\textbf{81.78}}{1.14}
& \stdcell{\textbf{57.34}}{1.93}
& \stdcell{93.51}{2.81}
& \stdcell{\textbf{87.79}}{1.44}
& \stdcell{\textbf{76.78}}{0.86}
& \stdcell{\textbf{0.169}}{0.009} \\
\midrule
\textbf{Dataset} & \textbf{Variant} & \textbf{OS Acc.\(\uparrow\)} & \textbf{RJ Acc.\(\uparrow\)} & \textbf{LC Acc.\(\uparrow\)} & \textbf{ES Acc.\(\uparrow\)} & \textbf{Avg. Acc.\(\uparrow\)} & \textbf{KL\(\downarrow\)} \\
\midrule
\multirow{2}{*}{\textbf{OmniMedVQA}} & w/o trans. & \stdcell{62.68}{1.23}
& \stdcell{78.81}{1.33}
& \stdcell{64.41}{1.80}
& \stdcell{\textbf{87.80}}{2.95}
& \stdcell{68.55}{0.80}
& \stdcell{0.359}{0.011} \\
\cmidrule(lr){2-8}
& \textbf{\method{} (ours)} & \stdcell{\textbf{64.54}}{1.21}
& \stdcell{\textbf{80.19}}{1.30}
& \stdcell{\textbf{70.20}}{1.72}
& \stdcell{82.11}{3.46}
& \stdcell{\textbf{70.83}}{0.79}
& \stdcell{\textbf{0.298}}{0.011} \\
\bottomrule
\end{tabular}
\end{table*}

Table~\ref{tab:frequency-ablation} isolates the transition prior at the main 10\% keep ratio by comparing \method{} with the corresponding non-transition control with \(\beta=0\). Adding the transition prior improves overall accuracy and most group accuracies, but the pattern is not uniform: SLAKE mainly gains in OT, while OmniMedVQA improves most in RJ and LC, where missed local structures are more likely to change the answer. The KL value changes little on OmniMedVQA, so this ablation is better read as an answer-level reranking benefit than as a general distribution-matching gain.

\subsection{Transition Weight Sensitivity}

\begin{figure}[t]
\centering
\includegraphics[width=0.8\linewidth]{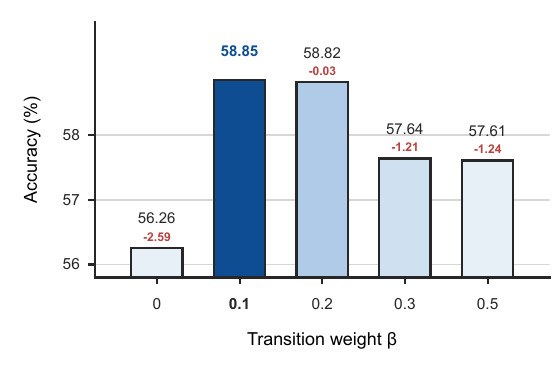}
\caption{Moderate transition weighting works best on OmniMedVQA at 10\% keep ratio with Lingshu-7B: \(\beta=0.1-0.3\) gives higher accuracy, while larger weights reduce accuracy.}
\label{fig:omnimed-beta-sensitivity}
\end{figure}

Figure~\ref{fig:omnimed-beta-sensitivity} studies the transition weight \(\beta\) on OmniMedVQA. Accuracy peaks at \(\beta=0.1-0.3\) and drops for larger weights, matching the view that the transition cue should refine the spectral ranking rather than dominate it. We therefore use it as a bounded correction to spectral support, not as an independent selector. \(\beta=0.2\) used as the default throughout our experiments.

\subsection{Efficiency Analysis}

We evaluate prefilling efficiency on SLAKE with Lingshu-7B using a single NVIDIA A100. Compared with the full-token baseline, \method{} at 10\% visual-token retention reduces the average LLM prefilling latency from 34.58 to 21.43\,ms, corresponding to a 38.03\% reduction and a \(1.61\times\) speedup. It also reduces incremental peak allocated GPU memory from 84.06 to 42.43\,MiB, a 49.52\% reduction. The average end-to-end latency decreases from 80.09 to 74.70\,ms, corresponding to a 6.74\% reduction and a \(1.07\times\) speedup.
\section{Discussion}

Across three benchmarks and two backbones, the results support that \method{} preserves diverse representation directions, while the bounded transition prior adjusts this selection toward localized feature changes. The pruning-rate and subspace-drift analyses are consistent with better representation preservation under aggressive compression. 

\paragraph{Limitations.}
The perturbation analysis motivates subspace preservation but does not guarantee answer correctness or prove that the discrete selector minimizes the spectral objective. Moreover, the transition prior is not a lesion detector: as a multiplier on complementary spectral contribution, it cannot recover a local cue with zero contribution to the current selection score. Our evaluation is also limited to two-dimensional medical VQA with two VLM backbones; extending the selector to volumetric images and broader clinical tasks remains future work.

\section{Conclusion}

We presented \fullmethod{}, a training free visual token pruning rule for medical VQA. \method{} combines spectral token influence, non redundant subset construction, and bounded token grid transition modulation. Experiments on SLAKE, OmniMedVQA and RadImageNet-VQA show improved fixed budget pruning under matched token budgets.

{
    \small
    \bibliographystyle{ieeenat_fullname}
    \bibliography{main}
}


\end{document}


\maketitle

\appendix
\section{Dataset and Group Details}
\label{sec:supp-data}

We evaluate on the same image-question instances used in the main paper: 2,394 SLAKE instances~\citep{liu2021slake}, 3,329 OmniMedVQA instances~\citep{hu2024omnimedvqa}, and 1,000 paired closed- and open-ended RadImageNet-VQA instances~\citep{butsanets2026radimagenetvqa}. The group labels are used only for reporting accuracy breakdowns; the pruning selector does not receive the group identity or imaging modality of an example. Table~\ref{tab:supp-dataset-groups} lists the group definitions and instance counts used throughout the main and supplementary tables.

\begin{table*}[t]
\centering
\footnotesize
\setlength{\tabcolsep}{4.1pt}
\caption{Dataset group and modality definitions with evaluated instance counts. Shares are computed within each dataset.}
\label{tab:supp-dataset-groups}
\begin{tabular}{@{}cllr@{}}
\toprule
\textbf{Dataset} & \textbf{Group / Modality} & \textbf{Meaning} & \textbf{Count / Share} \\
\midrule
\multirow{4}{*}{\textbf{SLAKE}}
& OR & Organ & 1,147 / 47.9\% \\
& AB & Abnormality & 654 / 27.3\% \\
& PO & Position & 77 / 3.2\% \\
& OT & Other & 516 / 21.6\% \\
\midrule
\multirow{4}{*}{\textbf{OmniMedVQA}}
& OS & Organ / structure & 1,458 / 43.8\% \\
& RJ & Radiology / joint / body region & 944 / 28.4\% \\
& LC & Lesion / cellular / tissue morphology & 804 / 24.2\% \\
& ES & Eye / skin & 123 / 3.7\% \\
\midrule
\multirow{2}{*}{\textbf{RadImageNet-VQA}}
& CT & Computed tomography images & 426 / 42.6\% \\
& MRI & Magnetic resonance imaging images & 574 / 57.4\% \\
\bottomrule
\end{tabular}
\end{table*}

SLAKE uses the official question-type grouping. OR questions focus on recognizing an organ, AB questions ask about abnormality, PO questions concern location or position, and OT covers the remaining question types. OmniMedVQA is grouped by semantic medical content so that the analysis separates anatomy-centric questions from radiology/body-region questions, lesion or tissue morphology questions, and eye/skin questions. The PO and ES groups are much smaller than the other groups, so their standard errors are naturally larger; we therefore interpret group-level changes together with the overall average accuracy and KL divergence. RadImageNet-VQA is grouped by image modality: CT and MRI.

\section{Additional Experimental Results}
\label{sec:supp-exp}

\subsection{Significance Analysis}

\begin{table*}[t]
\centering
\footnotesize
\setlength{\tabcolsep}{5pt}
\caption{Paired significance analysis. Confidence intervals are computed using
1,000 image-level paired cluster-bootstrap resamples. $p$-values are obtained
from two-sided exact McNemar tests.}
\label{tab:significance-analysis}
\begin{tabular}{@{}lllcc@{}}
\toprule
\textbf{Dataset}
& \textbf{Backbone}
& \textbf{Comparison}
& \textbf{95\% CI (pp)}
& \textbf{$p$-value} \\
\midrule
SLAKE
& Lingshu-7B
& MedSTP vs.\ BTP
& $[0.13,\,2.97]$
& $0.0375$ \\

SLAKE
& Qwen3-VL-8B
& MedSTP vs.\ VisPruner
& $[0.29,\,3.11]$
& $0.0175$ \\

OmniMedVQA
& Lingshu-7B
& MedSTP vs.\ PruneSID
& $[1.92,\,4.18]$
& $3.04{\times}10^{-7}$ \\

OmniMedVQA
& Qwen3-VL-8B
& MedSTP vs.\ BTP
& $[1.47,\,4.15]$
& $2.75{\times}10^{-5}$ \\

SLAKE
& Lingshu-7B
& MedSTP vs.\ w/o trans
& $[0.00,\,3.03]$
& $0.0436$ \\

OmniMedVQA
& Lingshu-7B
& MedSTP vs.\ w/o trans
& $[1.17,\,3.39]$
& $9.16{\times}10^{-5}$ \\

OmniMedVQA
& Qwen3-VL-8B
& MedSTP vs.\ w/o trans
& $[1.26,\,3.85]$
& $3.46{\times}10^{-4}$ \\

SLAKE
& Qwen3-VL-8B
& MedSTP vs.\ w/o trans
& $[1.90,\,4.55]$
& $3.30{\times}10^{-7}$ \\
\bottomrule
\end{tabular}
\end{table*}

Table~\ref{tab:significance-analysis} evaluates paired correctness differences on the same image--question instances. Against the strongest comparison method for each backbone and dataset, all four 95\% confidence intervals are positive and all exact McNemar tests give \(p<0.05\). 

\subsection{Transition-Weight Sensitivity}
\label{sec:supp-beta-sensitivity}

\begin{figure}[t]
    \centering
    \includegraphics[width=0.98\linewidth]{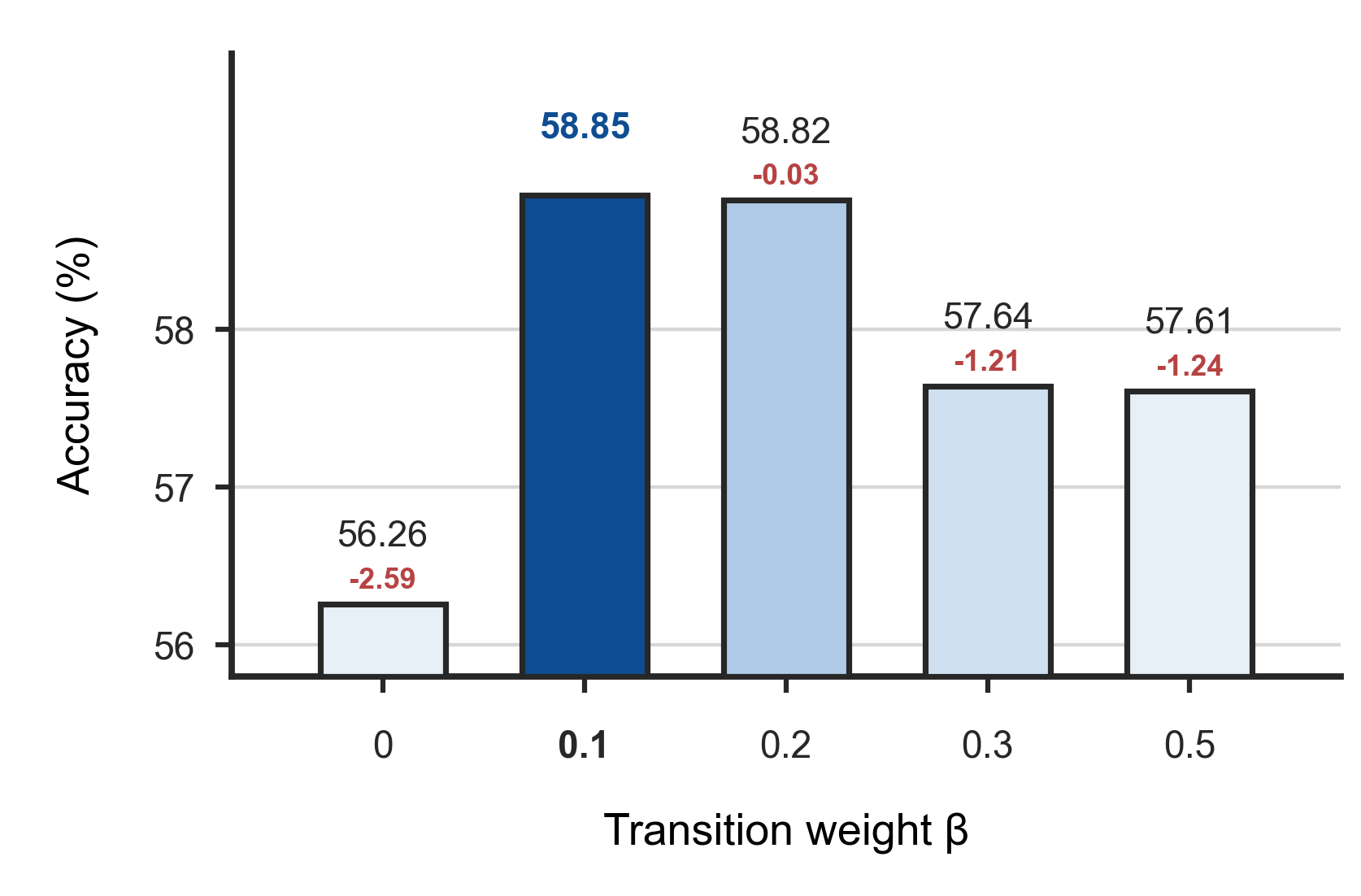}
    \caption{Transition-weight sensitivity on OmniMedVQA with Qwen3-VL-8B at a 10\% visual-token keep ratio. Accuracy peaks at \(\beta=0.1\), while the default \(\beta=0.2\) is within 0.03 percentage points; larger weights reduce accuracy. Red annotations show the difference from the best point estimate.}
    \label{fig:supp-beta-sensitivity}
\end{figure}

Figure~\ref{fig:supp-beta-sensitivity} studies the transition weight \(\beta\). Accuracy peaks at \(\beta=0.1-0.3\) and drops for larger weights, matching the view that the transition cue should refine the spectral ranking rather than dominate it. 

\subsection{Closest-Method Comparison at the 10\% Keep Ratio}

Table~\ref{tab:supp-closest} report the complete point estimates for paper-defined adaptations of ResPrune~\citep{li2026resprune}, SPARE~\citep{lee2026spare}, and SVD-Prune~\citep{apedo2026svdprune}. Official executable implementations were unavailable, so these rows should be interpreted as adaptations rather than official reproductions. All methods use a 10\% visual-token keep ratio.

\begin{table*}[t]
\centering
\footnotesize
\setlength{\tabcolsep}{3.5pt}

\caption{Closest-method comparison at a 10\% visual-token keep ratio. We report closed-ended accuracy (\%) on SLAKE, OmniMedVQA, and RadImageNet-VQA, together with open-ended RadImageNet-VQA accuracy (\%) evaluated by the method-blind Selene and Phi-4 semantic judges.}
\label{tab:supp-closest}

\begin{tabular}{@{}llcccccccc@{}}
\toprule

\multirow{3}{*}{\textbf{Backbone}}
& \multirow{3}{*}{\textbf{Method}}
& \multirow{3}{*}{\textbf{SLAKE}}
& \multirow{3}{*}{\textbf{OmniMedVQA}}
& \multicolumn{6}{c}{\textbf{RadImageNet-VQA}} \\
\cmidrule(lr){5-10}

& & & &
\multicolumn{3}{c}{\textbf{Closed-ended}}
& \multicolumn{3}{c}{\textbf{Open-ended}} \\
\cmidrule(lr){5-7}
\cmidrule(lr){8-10}

& & & &
\textbf{CT}
& \textbf{MRI}
& \textbf{Avg.}
& \textbf{Selene}
& \textbf{Phi-4}
& \textbf{Judge Avg.} \\

\midrule

\multirow{4}{*}{\textbf{Lingshu-7B}}
& ResPrune$^{\dagger}$
& \underline{68.05}
& 66.87
& 74.41
& \underline{69.69}
& \underline{72.05}
& 56.80
& 55.70
& 56.25 \\

& SPARE$^{\dagger}$
& 67.77
& \underline{67.17}
& \underline{75.12}
& 68.99
& \underline{72.05}
& \underline{57.70}
& \underline{56.20}
& \underline{56.95} \\

& SVD-Prune$^{\dagger}$
& 61.64
& 56.44
& 41.55
& 44.95
& 43.25
& 39.70
& 38.30
& 39.00 \\

& \textbf{\method{} (ours)}
& \textbf{76.78}
& \textbf{70.83}
& \textbf{79.81}
& \textbf{84.49}
& \textbf{82.15}
& \textbf{64.80}
& \textbf{62.40}
& \textbf{63.60} \\

\midrule

\multirow{4}{*}{\textbf{Qwen3-VL-8B}}
& ResPrune$^{\dagger}$
& \underline{54.38}
& \underline{52.90}
& \underline{78.17}
& \underline{85.54}
& \underline{81.85}
& 65.30
& \underline{60.70}
& \underline{63.00} \\

& SPARE$^{\dagger}$
& 53.91
& 52.30
& 76.29
& 82.23
& 79.26
& \underline{65.40}
& 60.40
& 62.90 \\

& SVD-Prune$^{\dagger}$
& 51.93
& 50.86
& 70.42
& 67.42
& 68.92
& 53.10
& 49.00
& 51.05 \\

& \textbf{\method{} (ours)}
& \textbf{78.45}
& \textbf{58.82}
& \textbf{83.10}
& \textbf{91.46}
& \textbf{87.28}
& \textbf{74.20}
& \textbf{69.90}
& \textbf{72.05} \\

\bottomrule
\end{tabular}

\vspace{1pt}

\parbox{0.98\textwidth}{
\raggedright\scriptsize
$^{\dagger}$Best-effort adaptation because official executable code was unavailable.
}

\end{table*}

\subsection{Qualitative Success and Failure Cases}
\label{sec:supp-qualitative-cases}

\begin{figure*}[t]
    \centering
    \includegraphics[width=0.98\textwidth]{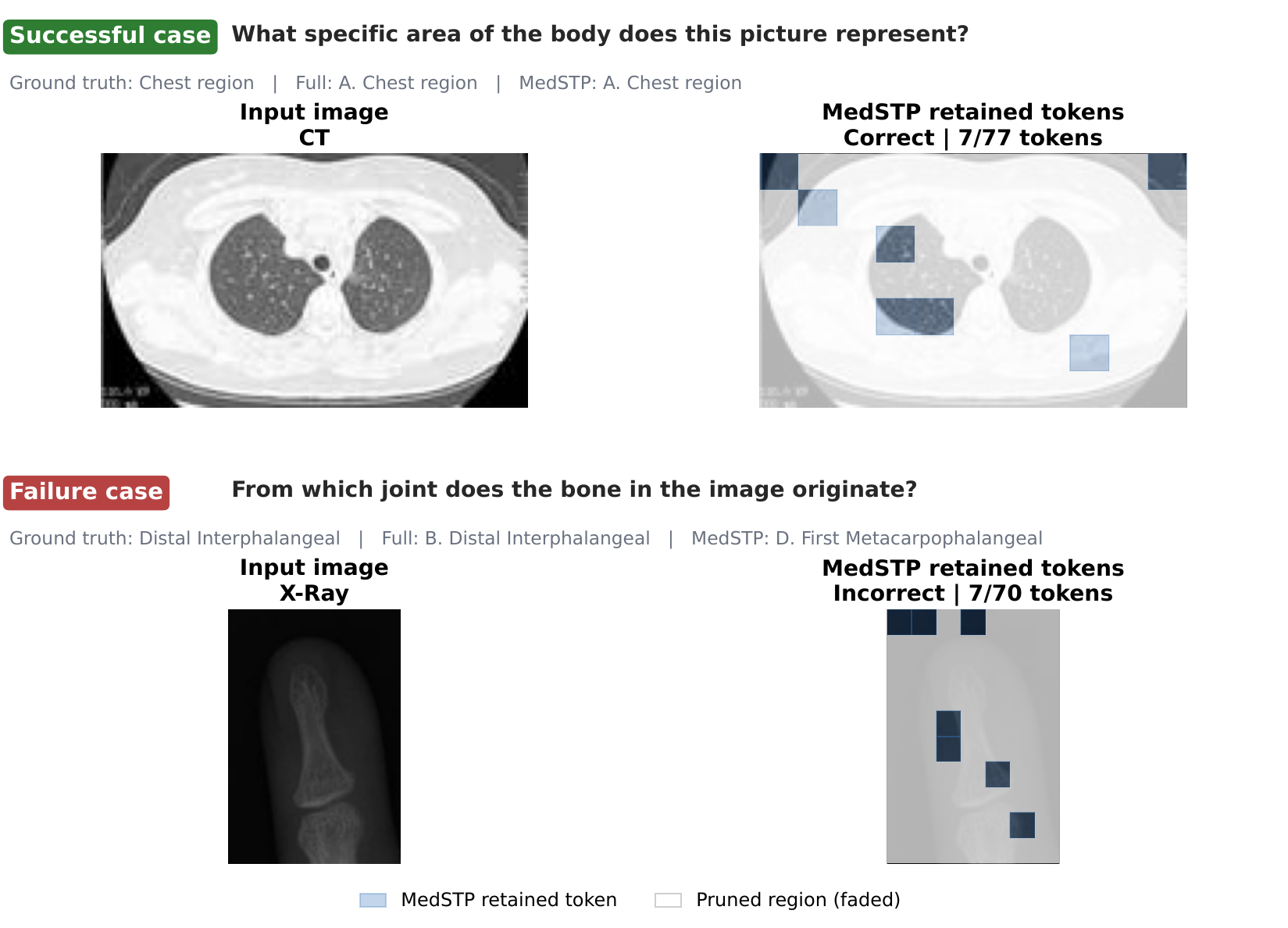}
    \caption{Qualitative success and failure cases on OmniMedVQA with Qwen3-VL-8B under the nominal 10\% visual-token keep setting. Blue cells mark the visual tokens retained by \method{} on the encoder-provided token grid; faded regions are pruned. }
    \label{fig:supp-omnimed-cases}
\end{figure*}

Figure~\ref{fig:supp-omnimed-cases} visualizes the actual visual-token indices retained by \method{} on two OmniMedVQA examples with Qwen3-VL-8B. In the successful CT example, the retained 7 of 77 visual tokens are sufficient to preserve the full-token prediction. In the X-ray example, the full-token model answers correctly, whereas pruning to 7 of 70 visual tokens changes the prediction to a different joint.

\clearpage

{
    \small
    \bibliographystyle{ieeenat_fullname}
    \bibliography{main}
}